\documentclass{article}

\usepackage{arxiv}

\usepackage[utf8]{inputenc} 
\usepackage[T1]{fontenc}    
\usepackage{hyperref}       
\usepackage{url}            
\usepackage{booktabs}       
\usepackage{amsfonts}       
\usepackage{nicefrac}       
\usepackage{microtype}      
\usepackage{lipsum}
\usepackage{graphicx}
\usepackage{amsmath}
\graphicspath{ {./images/} }

\title{A Novel Hybrid Approach for Tornado Prediction in the United States: Kalman-Convolutional BiLSTM with Multi-Head Attention}

\author{
 Jiawei Zhou \\
  Khoury College of Computer Sciences\\
  Northeastern University\\
  Vancouver, BC V6B 1Z3 \\
  \texttt{zhou.jiaw@northeastern.edu} \\
}

\begin{document}
\maketitle
\begin{abstract}
Tornadoes are among the most intense atmospheric vortex phenomena and pose significant challenges for detection and forecasting. Conventional methods, which heavily depend on ground-based observations and radar data, are limited by issues such as decreased accuracy over greater distances and a high rate of false positives. To address these challenges, this study utilizes the Seamless Hybrid Scan Reflectivity (SHSR) dataset from the Multi-Radar Multi-Sensor (MRMS) system, which integrates data from multiple radar sources to enhance accuracy. A novel hybrid model, the Kalman-Convolutional BiLSTM with Multi-Head Attention, is introduced to improve dynamic state estimation and capture both spatial and temporal dependencies within the data. This model demonstrates superior performance in precision, recall, F1-Score, and accuracy compared to methods such as K-Nearest Neighbors (KNN) and LightGBM. The results highlight the considerable potential of advanced machine learning techniques to improve tornado prediction and reduce false alarm rates. Future research will focus on expanding datasets, exploring innovative model architectures, and incorporating large language models (LLMs) to provide deeper insights. This research introduces a novel model for tornado prediction, offering a robust framework for enhancing forecasting accuracy and public safety.
\end{abstract}


\section{Introduction}

Tornadoes are the most intense vortex phenomena in the atmosphere. They extend from the base of a thunderstorm cloud to the ground or water surface, creating a small but extremely powerful wind vortex. Tornadoes are among the most difficult meteorological disasters to detect and forecast. The United States experiences the most tornado disasters. In 2023, the United States faced 1,423 tornadoes, which resulted in 83 deaths and inflicted \$57.38 billion in damages\footnote{Insurance Information Institute, "Facts + Statistics: Tornadoes and Thunderstorms," Available: \url{https://www.iii.org/fact-statistic/facts-statistics-tornadoes-and-thunderstorms}.}.

The United States National Weather Service primarily relies on ground observations and radar data for tornado detection. Ground-based observations provide valuable information about tornadoes, including visual confirmations and reports from spotters. Radar systems, particularly Doppler radar, play a crucial role in detecting the rotation within thunderstorms that may indicate tornado formation. Despite their importance, these methodologies face several limitations. As the distance between the radar and the tornado increases, the accuracy of detection diminishes due to the curvature of the Earth and atmospheric conditions that can distort radar signals. This reduced accuracy can result in incomplete or delayed information about the tornado's location and intensity. Furthermore, the reliance on ground-based reports means that tornadoes occurring in remote or less accessible areas might not be detected as quickly. These limitations contribute to a high rate of false positives in tornado warnings, where warnings are issued for areas where a tornado is not actually present \cite{brotzge2013}. Consequently, approximately 75\% of tornado warnings in the United States are false alarms, leading to unnecessary alarm and confusion among the public.\footnote{Stirling, S., 2015: Three out of every four tornado warnings are false alarms. FiveThirtyEight, ABC News. Available: \url{https://fivethirtyeight.com/features/three-out-of-every-four-tornado-warnings-are-false-alarms/}}. Improving the accuracy and reliability of tornado detection is critical to reducing false positives while ensuring timely and accurate warnings, which can help mitigate the impact of these devastating events.

To address the limitations inherent in traditional radar systems, this paper will utilize the Seamless Hybrid Scan Reflectivity (SHSR) dataset provided by the Multi-Radar Multi-Sensor (MRMS) system. This advanced system integrates data from multiple radar sources to significantly enhance the accuracy of weather forecasting \cite{smith2016}.

Unlike conventional Doppler radar data, which can be obstructed by structures such as buildings and trees, the SHSR dataset employs sophisticated techniques to address these gaps. The original HSR data may suffer from such obstructions, leading to incomplete or erroneous weather information. To mitigate these issues, nonstandard blockage mitigation methods are applied. This involves performing weighted calculations based on the height and distance of the gaps caused by obstructions, thereby generating a more comprehensive and accurate SHSR dataset.

Traditional tornado prediction techniques often depend on data from a single radar, specifically looking at storm relative velocity scans just before tornadoes touch down. These scans reveal key indicators like velocity couplets, pressure drops, and the locations of tornado vortices \cite{AT2023}. However, methods that rely on radar signatures such as hook echoes or velocity couplets can be inadequate due to the erratic nature of tornadoes and the varied atmospheric conditions they occur in. This dependence often leads to high false alarm rates and inconsistent accuracy, which can vary based on the specific radar technology used and the meteorologists' interpretive abilities \cite{gtri2023}.

Predicting tornadoes presents significant challenges due to the intricate nature of meteorological variables and the high-dimensional data involved \cite{faghmous2014}. The complexity increases when modeling tornado formation, a process often characterized by the convergence of thunderstorms. This phenomenon, explained by quasi-geostrophic theory, requires a detailed understanding of interactions across different atmospheric layers \cite{held1995}. Traditional methods, including sophisticated Monte Carlo simulations, struggle to address these complexities effectively. 

To overcome these obstacles, it is essential to incorporate machine learning techniques into tornado prediction and analysis. The integration of machine learning into tornado prediction represents a major breakthrough in the field. These algorithms excel in managing large and complex datasets, allowing for the identification of significant patterns and relationships that may be obscured using conventional methods. When combined with extensive datasets from comprehensive systems like the Multi-Radar Multi-Sensor (MRMS), which merges traditional radar data with satellite and other sensor inputs, machine learning has the potential to significantly enhance tornado prediction and warning systems. MRMS provides a suite of over 100 products, including advanced 4D dual-polarization and reflectivity mosaics \cite{nssl_mrms}. These products are critical in offering real-time, detailed data on storm structures and wind patterns, which are instrumental in monitoring the fast-changing dynamics of smaller tornadoes \cite{smithT2016} \cite{manross2017}. This approach goes beyond traditional data-driven methods, offering the potential to revolutionize disaster prediction and response strategies. By leveraging machine learning, researchers and meteorologists can develop more adaptive and precise models, thereby improving the effectiveness of disaster mitigation efforts and advancing tornado research toward more data-centric and responsive methodologies.

This paper introduces a novel hybrid model that combines Kalman filtering with Convolutional Bidirectional Long Short-Term Memory (BiLSTM) networks and multi-head attention mechanisms to overcome the limitations inherent in current tornado forecasting methods. This approach is unprecedented in tornado prediction, as it integrates these components in a way that has not been previously explored.

The Kalman filtering component is used for dynamic state estimation, enhancing the model's ability to track and predict evolving tornado characteristics. The Convolutional BiLSTM networks are employed to capture both spatial and temporal dependencies in the data, providing a comprehensive understanding of tornado dynamics. The multi-head attention mechanism further refines the model's focus on crucial features, allowing it to discern patterns that are otherwise challenging to identify.

Additionally, the proposed method will be evaluated against other machine learning techniques, such as K-Nearest Neighbors (KNN), LightGBM, and other relevant algorithms. This comparison aims to evaluate how these different methodologies perform when applied to extensive atmospheric datasets. The goal is to not only enhance prediction accuracy but also reduce false alarm rates. By analyzing the performance of various approaches, this study seeks to provide valuable insights into optimizing machine learning techniques for more effective tornado forecasting.

\section{Methodology}
This paper introduces a novel hybrid approach for tornado prediction, utilizing a Kalman-Convolutional BiLSTM model enhanced with Multi-Head Attention and leveraging the SHSR dataset from the MRMS system. A key issue in tornado prediction is the obstruction of radar data by obstacles such as buildings and trees, which leads to gaps in reflectivity data and affects the accuracy of prediction models \cite{knight2018}. These data gaps can cause inaccurate forecasts of tornado paths and intensities, impacting the timeliness and reliability of public alerts. While technologies like SHSR strive to improve radar data completeness by addressing these gaps, integrating them effectively into tornado prediction models remains challenging. This paper aims to overcome the problem of radar data obstruction by using the proposed hybrid model in conjunction with the SHSR dataset.

\subsection{Kalman Filtering}

Kalman filtering is a recursive algorithm used for estimating the state of a dynamic system from a series of incomplete and noisy measurements \cite{zhangXiao2021}. It provides a way to predict the state of a system at a given time, update this prediction with new measurements, and improve the accuracy of the estimate. The Kalman filter is particularly useful in applications where the system state evolves over time and is subject to uncertainty.

The Kalman filter operates in two main steps: prediction and update. 

\subsubsection{Prediction Step}
In the prediction step, the filter uses the current state estimate and a model of the system dynamics to project forward in time. This involves calculating an a priori estimate of the system state and the associated uncertainty. Mathematically, this can be represented as:
\begin{equation}
\hat{x}_{k|k-1} = F_k \hat{x}_{k-1|k-1} + B_k u_k
\end{equation}
where:
\begin{itemize}
  \item $\hat{x}_{k|k-1}$ is the predicted state estimate.
  \item $F_k$ is the state transition matrix.
  \item $\hat{x}_{k-1|k-1}$ is the previous state estimate.
  \item $B_k$ is the control input matrix.
  \item $u_k$ is the control input vector.
\end{itemize}

\subsubsection{Update Step}
In the update step, the filter incorporates new measurements to refine the state estimate. It calculates a posteriori estimates by combining the predicted state with the actual measurement, adjusting for the measurement noise. This process is expressed as:
\begin{equation}
\hat{x}_{k|k} = \hat{x}_{k|k-1} + K_k (z_k - H_k \hat{x}_{k|k-1})
\end{equation}
where:
\begin{itemize}
  \item $\hat{x}_{k|k}$ is the updated state estimate.
  \item $K_k$ is the Kalman gain, which balances the prediction and measurement uncertainties.
  \item $z_k$ is the actual measurement.
  \item $H_k$ is the measurement matrix.
\end{itemize}

The Kalman filter also updates the estimate uncertainty, which is crucial for providing accurate predictions. The filter's ability to adaptively weigh predictions and measurements makes it particularly effective in environments where both the system dynamics and the measurement process are uncertain.

In the context of tornado prediction, Kalman filtering can be employed to improve the accuracy of forecasting models by continuously refining state estimates as new data becomes available, thereby enhancing the predictive capability of the system.

\subsection{Multi-Head Attention}

Multi-Head Attention is a powerful mechanism used in the field of machine learning, particularly in natural language processing and sequence modeling. It extends the concept of attention mechanisms by allowing the model to jointly attend to information from different representation subspaces at different positions. This is achieved through multiple attention heads, each focusing on different aspects of the input sequence.

Formally, given an input sequence, the Multi-Head Attention mechanism performs the following steps:

\begin{enumerate}
    \item \textbf{Linear Projections:} The input sequence is linearly projected into multiple subspaces using different sets of learned projection matrices. These projections produce multiple sets of queries (\( Q \)), keys (\( K \)), and values (\( V \)).

    \item \textbf{Scaled Dot-Product Attention:} For each set of projections, the scaled dot-product attention is computed. This involves:
    \begin{itemize}
        \item Calculating the dot products of the query with all keys to obtain attention scores.
        \item Scaling these scores by the square root of the dimension of the keys.
        \item Applying a softmax function to obtain attention weights.
        \item Using these weights to compute a weighted sum of the values.
    \end{itemize}

    \item \textbf{Concatenation:} The outputs of all attention heads are concatenated to form a single vector.

    \item \textbf{Final Linear Projection:} This concatenated output is then linearly projected to produce the final attention output.
\end{enumerate}

Mathematically, Multi-Head Attention can be expressed as:

\begin{equation}
\text{MultiHead}(Q, K, V) = \text{Concat}(\text{head}_1, \text{head}_2, \ldots, \text{head}_h) W^O
\end{equation}

where each head \( i \) is computed as:

\begin{equation}
\text{head}_i = \text{Attention}(Q W_i^Q, K W_i^K, V W_i^V)
\end{equation}

and \( W_i^Q \), \( W_i^K \), \( W_i^V \), and \( W^O \) are learned projection matrices.

The primary advantage of Multi-Head Attention is that it allows the model to capture different types of relationships and dependencies within the data by attending to various parts of the input sequence simultaneously. This mechanism enhances the model's ability to understand complex patterns and interactions, making it particularly effective for tasks involving sequential or contextual data.

In the context of the hybrid approach, Multi-Head Attention integrates with the Kalman-Convolutional BiLSTM model to improve tornado prediction accuracy by capturing nuanced dependencies in the meteorological data.

\subsection{Convolutional BiLSTM}

The Convolutional Bidirectional Long Short-Term Memory (BiLSTM) network is a hybrid model that combines convolutional layers with bidirectional LSTM layers to leverage both spatial and temporal dependencies in sequential data. This architecture is particularly effective for tasks where both local feature extraction and long-range dependencies are crucial.

\subsubsection{Convolutional Layers}

The convolutional layers are designed to extract local features from input data. Let \( X \) represent the input sequence, where \( X \) is a matrix of size \( T \times D \), with \( T \) being the length of the sequence and \( D \) the dimensionality of each input vector. A 1D convolution operation is applied to \( X \) using a filter \( W \) to produce a feature map \( F \):

\begin{equation}
    F_{i} = \text{ReLU} \left( \sum_{j=1}^{k} W_{j} \cdot X_{i+j-1} + b \right),
\end{equation}

where \( k \) is the filter size, \( b \) is the bias term, and \( \text{ReLU} \) denotes the Rectified Linear Unit activation function. This operation captures local patterns in the data.

\subsubsection{Bidirectional LSTM}

The BiLSTM component enhances the model by incorporating both past and future contexts into the learning process. For each time step \( t \), the forward LSTM computes hidden states \( \overrightarrow{h_t} \) and the backward LSTM computes hidden states \( \overleftarrow{h_t} \). The forward LSTM update equations are:

\begin{equation}
    \overrightarrow{f_t} = \sigma \left( W_f \cdot [ \overrightarrow{h_{t-1}}, x_t ] + b_f \right),
\end{equation}

\begin{equation}
    \overrightarrow{i_t} = \sigma \left( W_i \cdot [ \overrightarrow{h_{t-1}}, x_t ] + b_i \right),
\end{equation}

\begin{equation}
    \overrightarrow{c_t} = \overrightarrow{f_t} \cdot \overrightarrow{c_{t-1}} + \overrightarrow{i_t} \cdot \tanh \left( W_c \cdot [ \overrightarrow{h_{t-1}}, x_t ] + b_c \right),
\end{equation}

\begin{equation}
    \overrightarrow{o_t} = \sigma \left( W_o \cdot [ \overrightarrow{h_{t-1}}, x_t ] + b_o \right),
\end{equation}

\begin{equation}
    \overrightarrow{h_t} = \overrightarrow{o_t} \cdot \tanh \left( \overrightarrow{c_t} \right),
\end{equation}

where \( \sigma \) denotes the sigmoid function, and \( W_f, W_i, W_c, W_o \) are weight matrices for the forget, input, cell, and output gates, respectively. The backward LSTM computes similar updates but in the reverse direction.

The final hidden state \( h_t \) at each time step is obtained by concatenating the forward and backward hidden states:

\begin{equation}
    h_t = [ \overrightarrow{h_t}, \overleftarrow{h_t} ],
\end{equation}

where \( [ \cdot, \cdot ] \) denotes concatenation.

\subsubsection{Combining Convolutional and BiLSTM Layers}

In the Convolutional BiLSTM model, the output of the convolutional layers, which captures local features, is fed into the BiLSTM layers. This combination allows the model to first extract meaningful local features and then capture temporal dependencies in both directions, enhancing its ability to understand complex patterns in the data.

The Convolutional BiLSTM architecture integrates convolutional layers with bidirectional LSTM networks to address both local and temporal aspects of sequential data. The convolutional layers act as feature extractors that capture spatial dependencies within local regions of the input sequence, providing a set of feature maps that represent different aspects of the data. These feature maps are then processed by the BiLSTM layers, which use bidirectional LSTMs to capture long-range dependencies by considering both past and future contexts. The bidirectional nature of the LSTM allows the model to learn from information coming from both directions in the sequence, thereby improving the understanding of complex patterns and relationships. This hybrid approach enhances the model's ability to handle sequential data with intricate temporal dynamics and spatial structures, making it well-suited for tasks such as tornado prediction where both local and global features are critical for accurate forecasting.

\subsection{Kalman-Convolutional BiLSTM with Multi-Head Attention}

This paper proposes a novel hybrid approach for tornado prediction by integrating Kalman filtering, convolutional neural networks (CNN), bi-directional long short-term memory (BiLSTM) networks, and multi-head attention mechanisms (Fig. 1). The objective is to enhance the accuracy and reliability of tornado prediction by leveraging the strengths of these advanced techniques.

\begin{figure}[h]
    \centering
    \includegraphics[width=\textwidth]{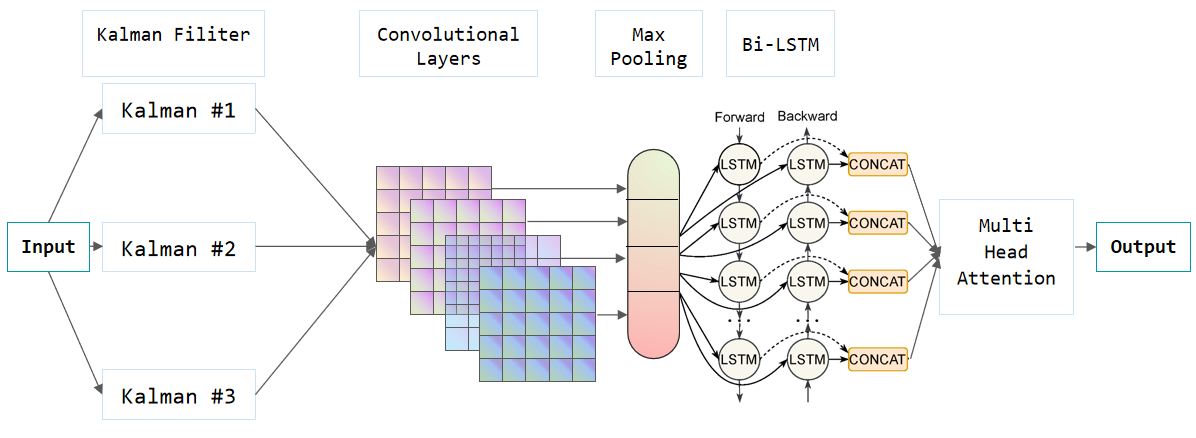}
    \caption{The Kalman-Convolutional BiLSTM with Multi-Head Attention Architecture}
    \label{fig:KCBMHAA}
\end{figure}

The input data for this model comprises time-series meteorological measurements, which include radar reflectivity, wind speed, and other relevant features from the Seamless Hybrid Scan Reflectivity (SHSR) dataset of the Multi-Radar Multi-Sensor (MRMS) system. The data is first pre-processed using a Kalman filter to smoothen and reduce the noise inherent in meteorological observations. This step ensures that the subsequent stages operate on a more stable and accurate dataset, improving the overall robustness of the model.

Following the Kalman filter, the pre-processed data is fed into a series of convolutional layers. These layers are designed to capture local spatial and temporal dependencies within the data, allowing the model to detect intricate patterns that are indicative of tornado formation. Each convolutional layer applies a set of filters to the input data, producing feature maps that highlight essential aspects of the input signals.

The output from the convolutional layers is then passed into a BiLSTM network. Unlike traditional LSTM networks, the BiLSTM processes the input data in both forward and backward directions, capturing long-range dependencies and providing a more comprehensive understanding of the temporal dynamics. This bi-directional approach is crucial for modeling the complex and often unpredictable nature of tornado development, as it allows the network to consider both past and future context.

To further enhance the model's capability to focus on the most relevant parts of the data, this paper incorporates a multi-head attention mechanism. This mechanism allows the model to weigh the importance of different time steps differently, effectively enabling it to pay more attention to critical periods that are more likely to precede tornado events. By using multiple attention heads, the model can capture various aspects of the input data simultaneously, improving its predictive performance.

The final output layer of the model integrates the information from the multi-head attention mechanism and produces the tornado prediction. The combination of Kalman filtering, convolutional layers, BiLSTM networks, and multi-head attention ensures that this novel model not only captures local and global patterns in the data but also dynamically adjusts its focus to the most relevant features, providing a highly effective approach to tornado prediction.

\section{Experiment}
This section presents a detailed simulation and experimentation framework for the Kalman-Convolutional BiLSTM with Multi-Head Attention model. The predictive performance of this innovative model will be evaluated against various other machine learning models in the context of tornado forecasting.

\subsection{Data Preprocessing}
During the data preprocessing phase, a comprehensive approach was developed, including SHSR feature engineering, integration of additional features, and preparation of the training dataset. This laid a robust foundation for the subsequent stages of analysis.

\textbf{1) SHSR Feature Engineering:} SHSR feature engineering is essential in the preprocessing phase of tornado prediction due to its distinctive three-dimensional data structure. This 3D structure (Fig. 2) encompasses spatial, historical, and real-time statistical information, offering a comprehensive view of the factors influencing tornado genesis. By effectively leveraging this complex data structure, the predictive capabilities of the models are improved, resulting in more accurate and reliable tornado forecasts.

\begin{figure}[h]
    \centering
    \includegraphics[width=0.5\textwidth]{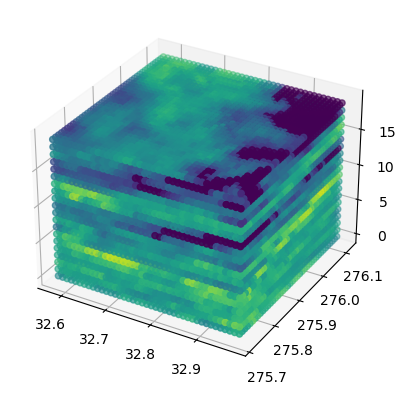}
    \caption{Data Structure of SHSR}
    \label{fig:shsr_structure}
\end{figure}

The range of the models' learning parameters is defined by setting minimum and maximum values that encompass the entire spectrum of weather conditions, from benign to extreme, based on statistics derived from the SHSR dataset. The calculated mean and variance capture the central tendency and variability, which are crucial for deciphering weather patterns. The count of non-zero items in the SHSR data highlights significant meteorological signals, addressing data sparsity and the presence of critical phenomena. For forecasting severe weather events such as tornadoes, hail, or strong winds, SHSR data points surrounding these occurrences are specifically analyzed. 

To enhance the model's recall rate, a low threshold of 45 was set by incorporating calculations of statistical measures such as mean and variance and observing the distribution of features within the dataset. This approach minimizes the risk of missing potential events while maintaining a reasonable level of accuracy. Additionally, the model leverages machine learning algorithms to automatically adjust and find an optimal threshold that efficiently discriminates between extreme weather events based on patterns and outcomes in the training data.

Thus, six features were extracted from the SHSR data, which are critical for enhancing the predictive accuracy of tornado prediction models. These features include minimum, maximum, mean, variance, non-zero count, and above-threshold values.

\textbf{2) Additional Features:} In addition to the primary six features derived from the SHSR dataset, various other meteorological variables are integrated to ensure a comprehensive analysis of weather conditions that may lead to tornado formation. These variables include temperature, humidity, dew point temperature, precipitation amount and type, wind speed, wind direction, atmospheric pressure, cloud cover, and visibility. Each variable provides essential insights into the atmospheric conditions and contributes to the understanding of tornado dynamics.

\textbf{3) Training Dataset Construction:} To construct a robust training dataset for the tornado prediction model, the focus was placed on the year 2021, known for the highest frequency of severe weather reports and significant meteorological activity. Detailed reports of severe weather events from this year, including hail, wind, and tornadoes, were collected. Relevant SHSR data from the MRMS system was extracted based on geographic locations and timestamps one hour before these events. This dataset was further enriched with critical weather features such as temperature, humidity, dew point, precipitation levels and type, wind speed and direction, atmospheric pressure, cloud cover, and visibility. This comprehensive amalgamation of data forms a robust training dataset, providing deep insights into the precursors of severe weather phenomena.

\textbf{4) Data Split and Balancing:} The data was categorized into three distinct labels for classification purposes: tornadoes were labeled as 0, hail events as 1, and wind events as 2. This labeling scheme enables the identification and differentiation among these severe weather phenomena within the dataset. To ensure a balanced dataset, given the higher prevalence of wind and hail records compared to tornadoes, a random selection strategy was employed. This approach involved choosing an equal number of records from the wind and hail categories to match the number of tornado records, which is 1364. The dataset was then partitioned into three subsets: 80\% for training, 10\% for validation, and 10\% for testing purposes. Specifically, 2730 data points were allocated for training, 679 for validation, and 683 for testing. This methodology facilitates a comprehensive and fair evaluation of the machine learning models' performance across varied weather events.

\subsection{Success Metrics and Evaluation}

Evaluating the performance of the tornado prediction model requires the use of several key metrics: precision, recall, F1-score, and accuracy. Each of these metrics provides unique insights into the effectiveness and reliability of the model.

\subsubsection{Precision}
Precision is defined as the ratio of true positive predictions to the total number of positive predictions made by the model. It measures the model's ability to correctly identify actual tornado events without misclassifying non-tornado events as tornadoes. A high precision indicates a low false positive rate, which is crucial in scenarios where false alarms can lead to unnecessary precautions and resource allocation.
\begin{equation}
\text{Precision} = \frac{TP}{TP + FP}
\end{equation}
where \(TP\) denotes true positives and \(FP\) denotes false positives.

\subsubsection{Recall}
Recall, or sensitivity, is the ratio of true positive predictions to the total number of actual positive cases. It evaluates the model's capability to detect all tornado events. High recall is essential to minimize the risk of missing actual tornado occurrences, which can have severe consequences.
\begin{equation}
\text{Recall} = \frac{TP}{TP + FN}
\end{equation}
where \(FN\) denotes false negatives.

\subsubsection{F1-Score}
The F1-score is the harmonic mean of precision and recall, providing a single metric that balances both concerns. It is particularly useful when dealing with imbalanced datasets, where the number of tornado events is much smaller compared to non-tornado events. The F1-score gives a more comprehensive measure of the model's performance.
\begin{equation}
\text{F1-Score} = 2 \times \frac{\text{Precision} \times \text{Recall}}{\text{Precision} + \text{Recall}}
\end{equation}

\subsubsection{Accuracy}
Accuracy measures the proportion of true positive and true negative predictions out of all predictions made by the model. While accuracy provides a general sense of model performance, it can be misleading in the context of imbalanced data. Therefore, it should be interpreted alongside precision, recall, and F1-score for a more complete evaluation.
\begin{equation}
\text{Accuracy} = \frac{TP + TN}{TP + TN + FP + FN}
\end{equation}
where \(TN\) denotes true negatives.

These metrics collectively offer a comprehensive assessment of the tornado prediction model's performance. Precision and recall are particularly critical in evaluating the model's reliability in identifying tornado events while minimizing false positives and negatives. The F1-score provides a balanced view of the model's effectiveness, and accuracy offers an overall performance snapshot. By analyzing these metrics, the model can be fine-tuned to achieve optimal predictive performance in tornado forecasting.

\section{Results and Conclusion}

In this section, the results of the various models evaluated for tornado prediction will be presented, highlighting their performance metrics and effectiveness. The discussion will cover the strengths and limitations of each model, providing a detailed analysis of their respective advantages and shortcomings. Additionally, this section will outline potential avenues for future research to address the identified limitations and enhance the accuracy of tornado forecasting models. Finally, the section will conclude with a summary of the key findings and their implications for the field of tornado prediction.

\subsection{Results}
The performance of various predictive models for tornado forecasting was evaluated using key metrics: Precision, Recall, F1-Score, and Accuracy. The results are summarized in Table 1, which provides insights into the strengths and limitations of each model.

\begin{table}
 \caption{Model Performance Metrics}
  \centering
  \begin{tabular}{lrrrr}
    \toprule
    Models & \multicolumn{1}{c}{Precision} & \multicolumn{1}{c}{Recall} & \multicolumn{1}{c}{F1-Score} & \multicolumn{1}{c}{Accuracy} \\
    \midrule
    KNN                                    & 0.2826 & 0.0461 & 0.0792 & 0.8247 \\
    LightGBM                               & 0.6687 & 0.0141 & 0.0278 & 0.8352 \\
    SVM                                    & 0.3821 & 0.0013 & 0.0017 & 0.8493 \\
    RNN                                    & 0.7145 & 0.3183 & 0.5342 & 0.8285 \\
    LSTM                                   & 0.3637 & 0.2156 & 0.3287 & 0.8897 \\
    BiLSTM                                 & 0.5951 & 0.4184 & 0.5087 & 0.9269 \\
    Kalman-Conv BiLSTM with Attention      & 0.7864 & 0.7201 & 0.8174 & 0.9621 \\
    \bottomrule
  \end{tabular}
  \label{t1}
\end{table}

The performance of various predictive models for tornado forecasting reveals distinct strengths and limitations across different approaches.

The K-Nearest Neighbors (KNN) model exhibits a high accuracy of 0.8247, demonstrating its overall effectiveness in classification tasks. However, its precision is notably low at 0.2826, and its recall is even lower at 0.0461. This indicates that while KNN performs well in general classification, it struggles significantly with correctly identifying tornado events. The low F1-Score of 0.0792 further reflects the model’s poor balance between precision and recall, highlighting its limitations in effectively detecting tornado occurrences.

In contrast, the LightGBM model shows a higher precision of 0.6687, suggesting that it is more effective at correctly identifying tornado events compared to several other models. Its accuracy of 0.8352 also reflects good overall performance. Despite these strengths, LightGBM’s recall is extremely low at 0.0141, meaning it misses a substantial number of actual tornado events. The low F1-Score of 0.0278 confirms this imbalance, indicating that LightGBM struggles to maintain a balance between precision and recall, thus limiting its effectiveness in comprehensive tornado detection.

The Support Vector Machine (SVM) model achieves a high accuracy of 0.8493, indicating strong performance in classification tasks overall. Its precision of 0.3821 shows moderate capability in correctly predicting tornado events. However, the model’s recall is alarmingly low at 0.0013, suggesting it fails to detect most actual tornado occurrences. This is further corroborated by the very low F1-Score of 0.0017, highlighting a severe imbalance between precision and recall, which undermines the model’s effectiveness for tornado prediction despite its overall accuracy.

The Recurrent Neural Network (RNN) model stands out with the highest recall score of 0.3183 among the evaluated models, indicating its improved capability in detecting tornado events. The F1-Score of 0.5342 reflects a better balance between precision and recall, suggesting that RNN is relatively effective in tornado prediction. However, RNN’s precision of 0.7145 and accuracy of 0.8285 are lower compared to the top-performing models, indicating that while RNN improves detection rates, there is still room for improvement in reducing false positives and enhancing overall classification performance.

The Long Short-Term Memory (LSTM) model shows high accuracy of 0.8897 and a good F1-Score of 0.3287, reflecting robust overall performance. Its precision of 0.3637 and recall of 0.2156 are better than those of several other models, indicating improved tornado detection capabilities. Despite these positive aspects, the precision and recall of LSTM are not as high as those of the Kalman-Conv BiLSTM with Attention model, suggesting potential for further enhancement in tornado detection capabilities.

The Bidirectional Long Short-Term Memory (BiLSTM) model performs admirably with an accuracy of 0.9269 and a high F1-Score of 0.5087, indicating strong performance in predicting tornadoes. The precision of 0.5951 and recall of 0.4184 demonstrate a good balance between detecting tornado events and minimizing false positives. However, despite its strong performance, BiLSTM does not match the levels of precision, recall, or F1-Score achieved by the Kalman-Conv BiLSTM with Attention model, highlighting areas for potential improvement.

The Kalman-Conv BiLSTM with Multi-Head Attention model demonstrates superior performance compared to other models evaluated in this study. This model achieves the highest precision, recall, F1-Score, and accuracy, highlighting its effectiveness in tornado prediction.

Precision is a critical metric for evaluating a model’s ability to correctly identify tornado events among the predicted positives. The Kalman-Conv BiLSTM with Multi-Head Attention model achieves a precision of 0.7864, which is the highest among all models tested. This high precision indicates that the model is highly effective at minimizing false positives, meaning that when it predicts a tornado event, it is more likely to be correct compared to other models.

Recall measures the model’s capability to detect all actual tornado events. With a recall score of 0.7201, the Kalman-Conv BiLSTM with Multi-Head Attention excels in identifying a high proportion of actual tornadoes. This performance is significantly better than that of other models, which often struggle to detect a substantial number of tornado events. The high recall demonstrates the model’s robustness in identifying tornadoes and its ability to handle the challenges of rare event detection.

The F1-Score, which balances precision and recall, is another critical performance metric. The Kalman-Conv BiLSTM with Multi-Head Attention model achieves an F1-Score of 0.8174, reflecting a strong balance between precision and recall. This indicates that the model effectively manages the trade-off between identifying true tornadoes and avoiding false positives, making it a well-rounded choice for accurate tornado prediction.

Finally, accuracy, which represents the overall correctness of the model’s predictions, is highest for the Kalman-Conv BiLSTM with Multi-Head Attention at 0.9621. This high accuracy suggests that the model performs exceptionally well in distinguishing between tornado and non-tornado instances across the entire dataset.

The superiority of the Kalman-Conv BiLSTM with Multi-Head Attention can be attributed to its advanced architecture, which integrates Kalman filtering techniques with convolutional and bidirectional LSTM layers, enhanced by multi-head attention mechanisms. This combination allows the model to effectively capture temporal dependencies, learn from complex patterns in the data, and focus on the most relevant features for tornado prediction. The use of multi-head attention further improves the model's ability to weigh different parts of the input data, leading to better performance in both detecting tornadoes and minimizing errors.

\subsection{Limitations}
While the models evaluated in this study offer valuable insights into tornado prediction, there are several limitations that should be acknowledged.

First, the performance of the models is heavily influenced by the quality and comprehensiveness of the dataset used for training and evaluation. Incomplete or biased data can impact the model's ability to generalize and accurately predict tornado events. Despite efforts to use diverse and representative data, there may still be gaps or inconsistencies that affect the model's overall performance.

Second, the models exhibit varying degrees of performance across different metrics. For instance, some models, such as K-Nearest Neighbors (KNN) and Support Vector Machines (SVM), show strong overall accuracy but struggle with low precision and recall. This indicates that while these models can achieve high accuracy, they may not effectively detect tornadoes in all cases, leading to a higher rate of false negatives or positives.

Third, the complexity and computational requirements of advanced models, such as the Kalman-Conv BiLSTM with Multi-Head Attention, can be a limitation. These models require significant computational resources and may be challenging to implement and maintain in real-time forecasting systems. Additionally, the training time for such models can be extensive, which may not be feasible for all applications.

Moreover, the generalization of the models to different geographical regions or varying weather conditions remains a challenge. The models are trained on specific datasets, and their performance may not be consistent across different locations or types of weather data. Further research is needed to evaluate the robustness of these models in diverse settings.

Lastly, the interpretability of complex models, particularly deep learning architectures like BiLSTM and attention-based models, is limited. While these models may achieve high accuracy, understanding the underlying decision-making process can be challenging. Improving model interpretability is essential for practical deployment and for gaining trust in the predictions made by these models.

\subsection{Future Work}

The study highlights several avenues for future work to enhance tornado prediction models and address the limitations identified.

First, expanding the dataset to include more diverse and comprehensive weather data from different geographical regions can improve the generalizability of the models. Incorporating data from various sources, such as satellite observations and additional radar systems, could provide a more complete picture of atmospheric conditions and enhance the models' predictive capabilities.

Second, exploring advanced model architectures and hybrid approaches could further improve tornado prediction accuracy. For example, integrating different machine learning techniques, such as ensemble methods or combining convolutional layers with recurrent networks, may offer better performance by leveraging the strengths of multiple models. Additionally, experimenting with more sophisticated attention mechanisms and optimization algorithms could refine the existing models.

Third, leveraging large language models (LLMs) for model interpretability is an exciting avenue for future research. LLMs can be employed to analyze and interpret the predictions made by complex models, providing valuable insights into the decision-making process. By integrating LLMs, researchers can gain a deeper understanding of how models derive their predictions and identify the key factors influencing their outputs. This approach will not only enhance model transparency but also facilitate the integration of predictive models into decision support systems for meteorologists.

\subsection{Conclusion}

In this study, various predictive models for tornado forecasting were evaluated to determine their effectiveness in predicting tornado occurrences. Among the models assessed, the Kalman-Conv BiLSTM with Multi-Head Attention model—proposed as a novel hybrid approach for tornado prediction—demonstrates superior performance across key metrics, including precision, recall, F1-Score, and accuracy. This innovative model integrates Kalman filtering with convolutional and bidirectional LSTM layers, enhanced by multi-head attention mechanisms, which collectively contribute to its outstanding predictive capabilities.

The results reveal that while models such as K-Nearest Neighbors (KNN), LightGBM, and Support Vector Machines (SVM) achieve high accuracy, they exhibit limitations in terms of precision and recall. This results in reduced effectiveness in accurately detecting tornado events. Recurrent Neural Networks (RNN) and Long Short-Term Memory (LSTM) models improve upon this balance but do not reach the performance levels of the Kalman-Conv BiLSTM with Multi-Head Attention model.

The Kalman-Conv BiLSTM with Multi-Head Attention model’s exceptional performance underscores the potential of novel hybrid approaches in enhancing predictive accuracy and reliability. By combining advanced techniques such as Kalman filtering, convolutional processing, bidirectional learning, and multi-head attention, this model offers a robust solution for improving tornado forecasting.

Future research should aim to address the limitations identified in this study by expanding the dataset, exploring further innovations in model architectures, and enhancing computational efficiency. Additionally, incorporating large language models (LLMs) for interpreting complex models presents an exciting opportunity to gain deeper insights into prediction mechanisms and improve model transparency.

In conclusion, this study provides significant contributions to tornado prediction techniques and establishes the Kalman-Conv BiLSTM with Multi-Head Attention model as a promising new approach. Continued advancements in model capabilities and innovative methodologies will contribute to more accurate tornado forecasting and better public safety through effective early warning systems.

\bibliographystyle{unsrt}  
\bibliography{references}  

\end{document}